\pdfoutput=1

\documentclass[11pt]{article}

\usepackage{acl}

\usepackage{times}
\usepackage{latexsym}

\usepackage[T1]{fontenc}

\usepackage{adjustbox}
\usepackage{blindtext}

\usepackage{soul}
\usepackage{amsmath}
\usepackage{amssymb}
\usepackage{amsfonts}
\usepackage{xcolor}
\usepackage[utf8]{inputenc}

\usepackage{tcolorbox}

\newtcolorbox{ananya}[1][]{colback=red!5!white,colframe=red!75!black,fonttitle=\bfseries\small,before upper={\small},title=ananya}

\newtcolorbox{emma}[1][]{colback=orange!5!white,colframe=orange!75!black,fonttitle=\bfseries\small,before upper={\small},title=emma}

\newtcolorbox{iz}[1][]{colback=blue!5!white,colframe=blue!75!black,fonttitle=\bfseries\small,before upper={\small},title=iz}

\newtcolorbox{tom}[1][]{colback=green!5!white,colframe=green!75!black,fonttitle=\bfseries\small,before upper={\small},title=tom}

\usepackage{microtype}

\usepackage{inconsolata}

\usepackage{microtype}

\usepackage{multirow}
\usepackage{subfigure}
\usepackage{colortbl}
\usepackage{booktabs}
\usepackage{graphicx}
\usepackage{cleveref}
\usepackage{makecell}

\definecolor{darkolivegreen}{rgb}{0.33, 0.42, 0.18}

\makeatletter
\newcommand{\printfnsymbol}[1]{%
  \textsuperscript{\@fnsymbol{#1}}%
}

\makeatother

\title{Just CHOP: Embarrassingly Simple LLM Compression}

\author{
    \textbf{Ananya Harsh Jha}$^{\ast}$,~\textbf{Tom Sherborne}$^{\diamond}$,~\textbf{Evan Pete Walsh}$^{\ast}$,  \\ \textbf{Dirk Groeneveld}$^{\ast}$,~\textbf{Emma Strubell}$^{\dagger\ast}$,~\textbf{Iz Beltagy}$^{\ast}$ \\
    $^{\ast}$Allen Institute for Artificial Intelligence \\
    $^{\diamond}$Institute for Language, Cognition and Computation, University of Edinburgh \\
    $^{\dagger}$Language Technologies Institute, Carnegie Mellon University \\
    \texttt{ananyaj@allenai.org} \\
}

\begin{document}
\maketitle
\begin{abstract}
Large language models (LLMs) enable unparalleled few- and zero-shot reasoning capabilities but at a high computational footprint. A growing assortment of methods for compression promises to reduce the computational burden of LLMs in deployment, but so far, only quantization approaches have been demonstrated to be effective for LLM compression while maintaining zero-shot performance. A critical step in the compression process, the pretrain-then-finetune paradigm, has largely been overlooked when adapting existing pruning strategies to LLMs or proposing new ones. In this work, we show that embarrassingly simple layer pruning coupled with an extended language model pretraining as the finetuning phase produces state-of-the-art results against structured and even semi-structured compression of models at a 7B scale while being more inference efficient. We call this method LayerChop, where we deterministically remove layers from a model followed by task-agnostic finetuning of the remaining weights by continued self-supervised pretraining. At this scale, we also show how distillation, which has been super effective in task-agnostic compression of smaller BERT-style models, becomes inefficient against our simple pruning technique.
\end{abstract}


\section{Introduction}
\label{section:introduction}

Large language models (LLMs) have demonstrated efficacy in a wide range of real-world tasks mainly due to their few- and zero-shot reasoning capabilities ~\cite{gpt-3}. However, these successes come at a high computational cost of training, and operational constraints on memory and throughput demand effective compression methods for LLM deployment. 

While a flurry of methods for compressing LLMs have emerged~\citep{wanda,shortenedLlama} since the shift from small task-specific models to transfer learning large pretrained models, the vast majority of compression methods have been developed for the BERT-style, non-autoregressive (encoder-only) pretrain-then-finetune paradigm~\citep{distill-bert,homotopic-distill}. This differs substantially from the autoregressive (decoder-only) models that enable zero-shot reasoning via in-context learning. Aside from model architecture, the most notable difference is in the availability and scale of training data~\citep{dolma}.

The majority of compression methods have been developed for the \textit{task-specific} setting~\citep{specialized-heads, structured-bert-pruning-qa, sixteen-heads}, where the compressed model has access to supervised training data for task-specific finetuning or selection of parameters; these approaches are fundamentally incompatible with compressing an LLM for general-purpose in-context learning. Methods have also been developed for \textit{task-agnostic} BERT-style LM compression~\citep{early-bert, layer-dropout}, where the goal is to remove model parameters while maintaining language modeling perplexity. While task-agnostic compression is more compatible with the current paradigm, most BERT-style methods assume that training on the entire pretraining corpus is feasible~\citep{distill-bert,early-bert,homotopic-distill}. This assumption is no longer valid for LLMs.

As a result, unlike their non-autoregressive predecessors, methods for compressing autoregressive LLMs have focused on reducing model size with no subsequent finetuning, even on task-agnostic pretraining data. Under this constrained setting, methods compressing model parameters and activations via quantization have shown the most success to date \citep{llm-int8}, and although distillation and pruning methods have the potential to provide orthogonal benefits alongside model quantization, those benefits still need to be realized.

In this work, we demonstrate the utility of task-agnostic finetuning for compression of autoregressive LLMs. We explore two simple structured pruning techniques followed by continued pretraining under a limited token budget: LayerCHOP and DimCHOP. LayerCHOP reduces model depth by deterministically removing entire model layers. DimCHOP reduces model width by uniformly removing the same number of dimensions from each weight matrix in the network by adapting a parameter-level pruning criterion proposed in PLATON~\citep{platon} to dimensions. Furthermore, under the relaxed assumption that the compression procedure may include additional gradient updates on the pretraining data, we demonstrate that simply pretraining a smaller model from scratch on a limited budget of tokens represents a strong baseline against which we can compare compression methods. Finally, inspired by the success of augmenting task-agnostic compression with distillation objectives~\citep{homotopic-distill} we test this hypothesis for LLMs in the new pretrain-then-finetune paradigm.

Evaluation on language model perplexity and on a suite of 6 zero-shot tasks across three model sizes, we demonstrate the following:

\begin{itemize}
    \setlength{\itemsep}{0pt}
    \setlength{\parskip}{0pt}
    \setlength{\topsep}{0pt}
    \item LayerCHOP coupled with a finetuning phase, outperforms existing structured and even semi-structured compression at the 7B scale.
    \item Pretraining smaller models from scratch forms a strong baseline against which structured compression with a finetuning phase can be compared against.
    \item Both LayerCHOP and DimCHOP also outperform their distillation objective augmented variants to show that distillation does not scale efficiently in large model and data regimes.
\end{itemize}


\section{Related Work}
\label{section:related_work}

\paragraph{Pruning} seeks to identify sub-networks within larger models that yield 
the best possible performance with increased computational efficiency. Pruning uses heuristics~\citep{movement-pruning, differential-pruning} to identify parts of a model to prune.

Lottery ticket hypothesis~\citep{lth} introduces the concept of sparse sub-networks within a model, called winning tickets, which have the full model's task capability, but pruning of weights is unstructured. A different line of work introduces semi-structured pruning~\citep{structured-n-m-sparsity}, which improves throughput speed over unstructured pruning via accelerator-specific kernels. Structured pruning approaches~\citep{layer-dropout, structured-n-m-sparsity} aim to prune larger blocks within a model while considering architecture nuances for improving inference speed. Structured pruning can be coarse- or fine-grained. Coarse-grained pruning removes entire model layers~\citep{layer-dropout}. In the case of LMs, fine-grained pruning methods prunes atomic components like attention-heads~\citep{specialized-heads, sixteen-heads, differentiable-head-pruning} or hidden layers~\citep{dyna-bert, early-bert, structured-bert-pruning-qa}.

Another classification approach for pruning is based on task-agnostic vs. task-specific pruning. Task-agnostic pruning approaches~\citep{early-bert, layer-dropout} prune a model on pretraining data and then add task specialization as a second step. Task-specific pruning~\citep{specialized-heads, structured-bert-pruning-qa, sixteen-heads} methods specialize their models on end-task data during the pruning process.

More recently, pruning has been extended to larger decoder-only LLMs with semi-/un-structured~\citep{wanda,sparseGPT} and structured pruning methods~\citep{llmSurgeon,sliceGPT,shortenedLlama,sheared-llama}.

\paragraph{Knowledge Distillation} can be similarly divided between task-specific~\citep{patient-distill, pretrain-distill, xtreme-distill-transformer, Tang2019DistillingTK, co-fi} and task-agnostic methods~\citep{distill-bert, tiny-bert, mobile-bert, mini-lm, mini-lm-v2, homotopic-distill} for BERT-style models. Both task-specific and task-agnostic distillation methods finetune BERT-style encoder models on end tasks to evaluate them. Seq-KD~\citep{seq-KD} proposed task-agnostic distillation with a joint word level and sequence level objective for seq2seq models~\citep{attention-all-you-need}, which has been expanded to larger models like Distil Whisper~\citep{distil-whisper}.

Some methods follow pruning with a distillation phase~\citep{dyna-bert, structured-bert-pruning-qa}. More recently, methods like Co-Fi~\citep{co-fi} and Homotopic-distillation~\citep{homotopic-distill} combine pruning with distillation by applying distillation losses during the pruning process.


\begin{table*}[ht]
\centering
\small
\begin{tabular}{lp{0.3\linewidth}p{0.35\linewidth}l}
\toprule
\textbf{Method}            & \textbf{Encoder-only} & \textbf{Decoder-only} & \textbf{This work}             \\ \midrule
\textbf{Struct. depth}         & Poor Man's BERT \citep{poor-bert}   & ShortenedLlama~\citep{shortenedLlama}     & LayerCHOP                 \\
\textbf{Struct. width}     & PLATON \citep{platon}        & \makecell{SliceGPT \citep{sliceGPT};\\ LLMSurgeon \citep{llmSurgeon}}  & DimCHOP                   \\
\textbf{Un-/Semi-struct.} & LTH-BERT~\citep{lth-bert}         & \makecell{Wanda~\citep{wanda};\\ SparseGPT~\citep{sparseGPT}}    & ---                       \\
\textbf{Distillation}        & Homo-Distil~\citep{homotopic-distill}  & ---      & CHOP+distill \\
\midrule
\textbf{Training data} & Pretrain. (GBs) & Few batches & Pretrain. (TBs) \\ \bottomrule
\end{tabular}
\caption{Taxonomy of compression methods. We adapt pruning strategies from BERT-style models to LLMs under the pretrain-then-finetune paradigm. We also provide a list of recent or concurrent works for LLMs which compress them but do not follow it with a finetuning phase. We do not consider an un- or semi-structured form of CHOP as our early experiments validated our structured methods as having sufficient end-task accuracy while maintaining superior inference efficiency.}
\label{tab:taxonomy}
\end{table*}

\section{How to Train Your (Compressed) LLM}
\label{sec:training_compressed_llms}

This section describes methods for task-agnostic model compression of large language models (LLMs). From previous literature, we adapt pruning strategies that are simple, efficient, and go well with continued pretraining on a large data corpus described in Section~\ref{sec:experimental-setup}.

We prune models in two simple ways: model depth (LayerCHOP; \S\ref{sec:layer-chop}) and model width (DimCHOP; \S\ref{sec:dim-chop}). It has been observed in the BERT model compression literature that adding distillation in conjunction with pruning amplifies the result of compression~\citep{homotopic-distill}. Hence, we experiment with the same for decoder-only LLMs in the large data regime. A taxonomy of our adapted methods and where they lie within the current compression literature is summarized in Table~\ref{tab:taxonomy}.

\subsection{LayerCHOP}
\label{sec:layer-chop}

LayerCHOP is a layer removal pruning strategy for decoder-only transformer models for reducing model depth. We remove 50\% of layers from the pretrained Transformer architecture, corresponding to removing $\sim~50$\% of parameters. We continue (i.e., resume) training each model with a language modeling loss on data sampled from a similar corpus to the respective model's pretraining corpus (\S\ref{sec:experimental-setup}). Choosing where and when to prune layers is a critical design decision to maximize performance recovery after pruning. We experiment with five configurations for layer pruning shown in Figure~\ref{fig:pruning-locations} in Appendix~\ref{sec:appendix-layer-chop}. We always retain the first and the last layers as they interact with the embedding table and the final feed-forward layer of the model. Considering when to enact pruning primarily concerns either pruning all layers at initialization or incrementally removing layers periodically during the continued training of the model.

Our layer-pruning strategy is similar to that in Poor Man's BERT~\citep{poor-bert}. However, aside from the architectural differences between models (encoder-only vs decoder-only), our work contrasts in that we compress and evaluate in a stricter task-agnostic zero-shot setting without access to task-specific tuning. LayerDrop~\citep{layer-dropout}, similarly proposes \textit{layer dropout} as layer removal for compression. However, a key difference to our work is that layer dropout critically requires training from scratch with randomly masked layers to facilitate successful inference after pruning. Our method removes this constraint for broader utility---our technique can be applied to adapt and compress any publicly available LLM checkpoint. More recently, Shortened Llama~\citep{shortenedLlama}, has proposed layer pruning and then training LoRA~\citep{lora} weights on top of the pruned model. The main difference in their work is that they do not train their LoRA parameters at a pretraining corpus scale.

\subsection{DimCHOP}
\label{sec:dim-chop}

Our second method is defined for structured compression of decoder-only models in their width, i.e., dimensions, and we call it DimCHOP. This method is adapted for decoder-only LLM models from PLATON~\citep{platon}, a technique proposed for parameter pruning of encoder-only BERT models. For this method, we retain the same number of layers in the model as the original uncompressed version, but each layer has some dimensions with weights set to zero, e.g., for some layer $W\in\mathbb{R}^{m \times n}$ we apply mask $M\in\mathbb{R}^{m \times n}$, where $M_{n}=0$ and we take the element-wise Hadamard product $W \odot M$. We iteratively mask dimensions in weight matrices across the model in a balanced fashion, i.e., we zero out the same number of dimensions in each weight matrix of the model. This iterative pruning of dimensions is carried out as we continue training the model on language modeling loss on data sampled from a corpus similar to the pretraining set (\S\ref{sec:experimental-setup}).

The pruning criterion defined for selecting dimensions follows PLATON's sensitivity score $\mathcal{I}_j$ for each model parameter $j$ at some time during training $t$. Score $\mathcal{I}_j^t$ is the magnitude of the gradient-weight product (Eq.~\ref{eq:sensitivity-score}). PLATON further introduces an uncertainty score $\mathcal{U}_j$ (Eq.~\ref{eq:uncertainty}) for each parameter, representing the absolute difference between local sensitivity and an exponential weighted average parameter sensitivity (Eq.~\ref{eq:exp-sensitivity}). The final importance score (Eq.~\ref{eq:platon-criteria}) is the Hadamard product between the exponential moving averages of the sensitivity and the uncertainty scores.

\begin{align}
    \small
    \mathcal{I}_j^t &= |\theta_j^T \nabla \mathcal{L(\theta)}| \label{eq:sensitivity-score} \\
    \small
    \mathcal{\bar{I}}_j^t &= \beta_1 \mathcal{\bar{I}}_j^{t-1} + (1 - \beta_1) \mathcal{\bar{I}}_j^t \label{eq:exp-sensitivity}\\
    \small
    \mathcal{\bar{U}}_j^t &= |\mathcal{I}_j^t - \mathcal{\bar{I}}_j^t| \label{eq:uncertainty}\\
    \small
    \mathcal{\bar{U}}_j^t &= \beta_2 \mathcal{\bar{U}}_j^{t-1} + (1 - \beta_2) \mathcal{\bar{U}}_j^t \label{eq:exp-uncertainty}\\
    \small
    \mathcal{S}_t &= \mathcal{\bar{I}}_j^t \odot \mathcal{\bar{U}}_j^t\label{eq:platon-criteria}
\end{align}

While PLATON limits the importance scores to each parameter, we compute the importance scores for a dimension $\mathcal{I}_d$ (Eq.~\ref{eq:imp-score-dim}) by taking an $L_{1}$ norm over the importance scores of the parameters in the column of the weight matrix. The importance scores of all dimensions within a weight matrix are then sorted and the dimensions with the worst $k\%$ of the importance scores are set to 0 using the mask $M$. The $k\%$ of dimensions being set to 0 is defined by an exponential schedule which increases the number of dimension being pruned as the training progresses.

\begin{align}
    \small
    \mathcal{\bar{S}}_d^t &= ||\mathcal{\bar{S}}_{[:, j]}^t||_1
    \label{eq:imp-score-dim}
\end{align}

The difference between PLATON and our work is that PLATON prunes BERT-style models on specific end-task data at a parameter level, whereas we extend the idea of pruning parameters to pruning dimensions for decoder-only LLMs in a task-agnostic setting. This is more similar to Homotopic-Distillation~\citep{homotopic-distill}, which prunes dimensions from a BERT-style model, but augments the pruning with distillation objectives.

\subsection{Distillation-augmented Pruning}
\label{sec:distil-chop}

Homotopic distillation \citet{homotopic-distill} uses the PLATON pruning criterion discussed in \S\ref{sec:dim-chop}. This technique aggregates the importance score over weight matrix columns to prune an equal number of model dimensions from each weight matrix in a model. \citet{homotopic-distill} identifies 
that augmenting iterative dimension pruning with distillation-based continued training produces better compression outcomes in a task-agnostic setting. We evaluate a similar hypothesis in our structured compression setup of decoder-only models evaluated in a zero-shot fashion, and augment LayerCHOP and DimCHOP with distillation objectives. Eq.~\ref{eq:distillation:homotopic:total} gives the combined distillation objective for this setup comprising of the language modeling objective; KL-divergence over the outputs of the teacher and student; and mean-squared error distillation objective over attention outputs, intermediate layer representations, and the embedding table.

\begin{align}
    \small
    \mathcal{L}_{\Sigma} = \mathcal{L}_{\text{LM}} &+ \mathcal{L}_{\text{distill}} + \mathcal{L}_{\text{hid}} + \mathcal{L}_{\text{att}} + \mathcal{L}_{\text{emb}} \label{eq:distillation:homotopic:total} \\
    \small
    \mathcal{L}_{\text{hid}}(\theta_s, \theta_t) &= \sum_{k = 1}^K || H_t^k, H_s^k W_{\text{hid}}^k||_2^2 \label{eq:homotopic:hidden} \\
    \small
    \mathcal{L}_{\text{attn}}(\theta_s, \theta_t) &= \sum_{k = 1}^K || A_t^k, A_s^k||_2^2 \label{eq:homotopic:attention} \\
    \small
    \mathcal{L}_{\text{emb}}(\theta_s, \theta_t) &= || E_t, E_s W_{\text{emb}}||_2^2 \label{eq:homotopic:emb}
\end{align}

Eq.~\ref{eq:homotopic:hidden} defines the alignment between the intermediate representations of the model at each layer. We introduce a learned linear projection, $W^k_{\text{hid}}\in\mathbb{R}^{t\times s}$, to match the dimension between student and teacher representations. Eq.~\ref{eq:homotopic:attention} defines the mean squared error between attention outputs at each layer. Eq.~\ref{eq:homotopic:emb} defines the same loss between the embedding tables of the student and teacher models, with a learnable weight matrix $W_{\text{emb}}\in\mathbb{R}^{s_{\text{emb}}\times t_{\text{emb}}}$ matching dimensions between embeddings.


\section{Setup}
\label{sec:exp_setup}

\subsection{Experimental setup}
\label{sec:experimental-setup}

All our experiments use the C4~\citep{t5-c4} dataset for pretraining or pruning of models except for the pretrained llama2-7B~\citep{llama2}. C4 consists of approx. 160B web-crawled tokens. Our best-performing and most efficient compression strategy, LayerCHOP (Section ~\ref{sec:layer-chop}), trains on 1 complete epoch of C4, i.e., token budget of 160B tokens, for models at all scales. Other ablations (Section~\ref{sec:dim-chop}, ~\ref{sec:distil-chop}) train on smaller token budgets from the C4 training set. For language modeling evaluation, each experiment in the paper uses at least 13.2M C4 tokens from a separate validation set as we use 6400 batches at the full sequence length of the models used.

Our experiments use models at 3 scales: 300M, 1.1B, and 7B. For our 7B model pruning experiments, we use llama2-7B~\citep{llama2}. For the other two sizes, we train our own models on 1 epoch of C4 training set with 300M and 1.1B parameters. These smaller models follow PaLM~\citep{palm} like architecture with parallel attention and feed-forward network (FFN) blocks, SwiGLU activation~\citep{swiglu}, and fused FFN layers. All models in our experiment use flash-attention~\citep{flash-attn} for efficiency. We also train half-sized models for each scale, with half the number of layers as the base model, for pretraining comparisons in Section~\ref{sec:results-pretraining}. The architecture choices are summarized in Table~\ref{table:pretrained-model-configs} in Appendix~\ref{sec:appendix-arch}.

For experiments with the 300M and 1.1B models, we use the Lion optimizer \citep{lion-optim} with $\beta$ set to $\left(0.9, 0.95\right)$. Weight decay is $1 \times10^{-4}$ but is omitted for bias and normalization parameters. We found some stability issues with Lion at larger model sizes and reverted back to the more common AdamW~\citep{adamw} for our llama2-7B runs. The values of $\beta$ parameters are the same, while the weight decay is $0.1$. Each pruning and distillation experiment uses the same learning rate as pretraining, which is mentioned in Table~\ref{table:pretrained-model-configs} in Appendix~\ref{sec:appendix-arch}. The learning rate is warmed up to this value in 2000 steps and then decayed using a cosine schedule to 0.1 times the peak value. For our compression experiments, we start with a model checkpoint and train 500 steps before any compression is applied to a model to accumulate optimizer states.

\subsection{Evaluation setup}
\label{sec:eval-setup}

We evaluate our model on 6 tasks from 2 categories: common sense reasoning and science question answering. All tasks are evaluated in a zero-shot setting by providing the language model with a prompt from Eleuther-AI evaluation harness~\citep{eai-eval-harness} and a possible completion. We score the model output for each completion. The completion with the highest likelihood is the prediction to compute task accuracy. The completion likelihood can be normalized by the character count in the completion  \citep[length normalized accuracy]{eai-eval-harness}. Table~\ref{table:end-tasks} in Appendix~\ref{sec:appendix-eval} lists tasks in each evaluation category and respective metrics.


\section{Compression Results}
\label{sec:results}

The results section makes comparisons between task performance and efficiency of compression and inference for different variants of task-agnostic model compression. All results use the best configuration of LayerCHOP(Appendix.~\ref{sec:appendix-layer-chop}) and DimCHOP which is discussed in detail in the Appendix.

\subsection{Comparison against structured and unstructured pruning methods}

\begin{figure}[!bh]
\centering
\makebox[\columnwidth][c]{
\includegraphics[width=\columnwidth]{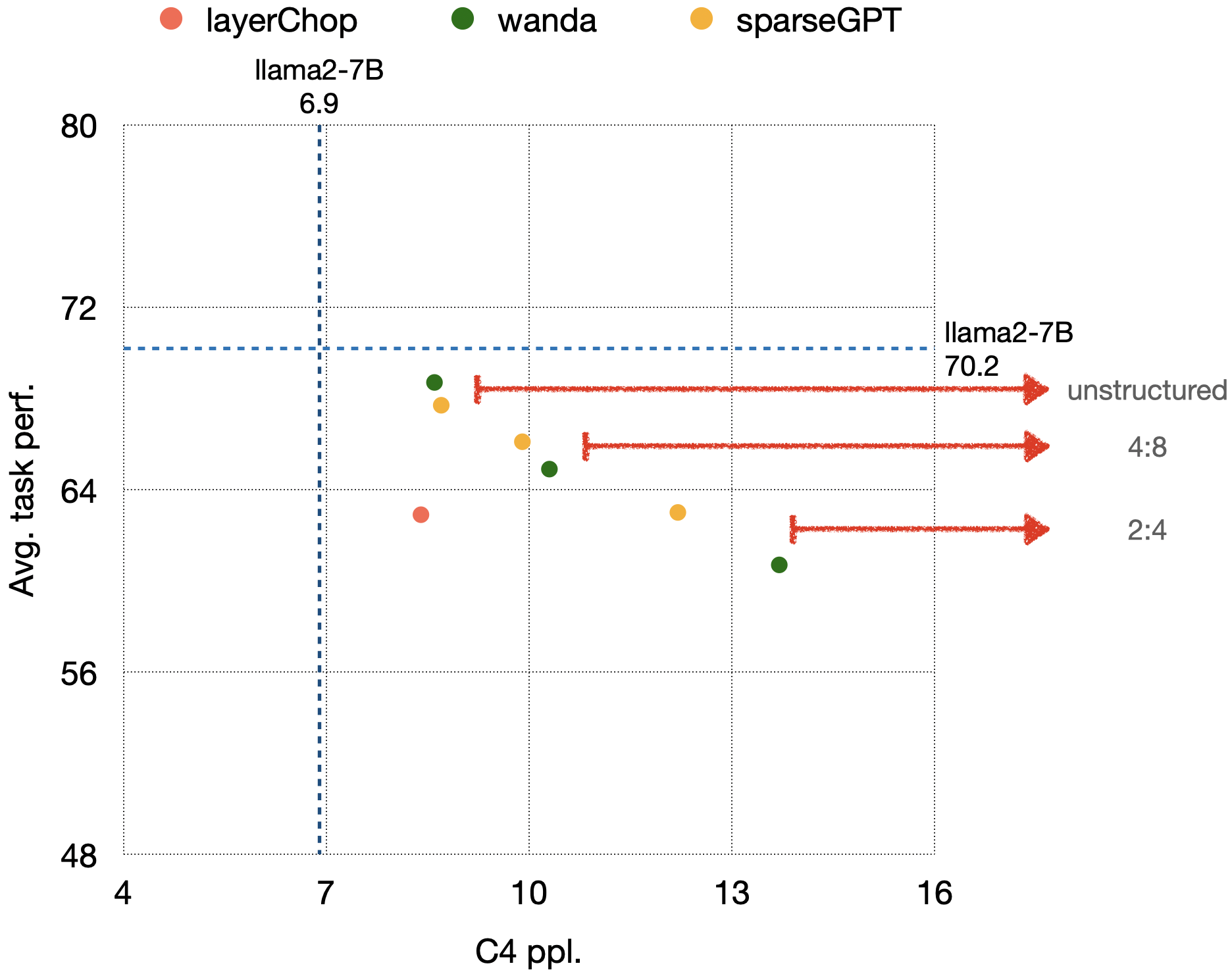}
}
\caption{Comparison between LayerCHOP and un-/semi-structured compression of llama2-7B from Wanda~\citep{wanda} and SparseGPT~\citep{sparseGPT}. Both Wanda and SparseGPT make their models public and for fair comparison we evaluate them on end tasks in our eval setup. We can see layerCHOP being better than 2:4 semi-structured pruned models and only marginally worse than 4:8 pruned models.}
\label{fig:struct-vs-unstruct}
\end{figure}

This section compares the results of LayerChop to concurrent unstructured/semi-structured and more recently proposed structured compression methods for LLMs. The other techniques we compare here do not continue pretraining their models on data after pruning. When combined with continued pretraining, we show how the simple pruning strategy of layer removal outperforms complicated pruning heuristics from other methods.

\begin{table*}[!t]
\centering
\footnotesize
\begin{tabular}{@{}c|ccc@{}}
\toprule
method                                                  & \textbf{shortenedLlama} & \textbf{sliceGPT} & \textbf{LLMSurgeon} \\ \midrule
\textbf{\# common eval tasks}                           & 5                        & 4                 & 5                   \\
\textbf{base model}                                     & llama-7B                 & llama2-7B         & llama2-7B           \\
\midrule
\textbf{method base avg./method pruned avg.}            & 68.9/57.9                & 74.7/56.6         & 68.6/46.6           \\
\textbf{\% model pruned/\% task score decrease} & 35\%/-16\% & 30\%/-24.2\% & 50\%/-32.1\% \\
\midrule
\textbf{our base avg./our pruned avg.}                  & 66.2/58.2                 & 72.0/62.8         & 66.2/58.2          \\
\textbf{our \% model pruned/our \% task score decrease} & 50\%/\textbf{-12.1\%}                & 50\%/\textbf{-12.8\%}     & 50\%/\textbf{-12.1\%}           \\ \bottomrule
\end{tabular}
\caption{Comparison against the concurrent works of Shortened Llama~\citep{shortenedLlama}, SliceGPT~\citep{sliceGPT}, and LLMSurgeon~\citep{llmSurgeon}. Since each paper has its own eval suite and base model, and not all of these methods have publicly available checkpoints, we compare them by averaging the base model scores and pruned model scores on end tasks which are common with our method. Then we find the percentage of end tasks score decrease with the percentage of model being pruned and list them in this table. As we can, LayerCHOP prunes the model to $50\%$ of the size, better than the other methods other than LLMSurgeon, and has the least reduction in average end task score across all methods.}
\label{tab:structured-comparison}
\end{table*}

In Figure~\ref{fig:struct-vs-unstruct}, we compare LayerChop to some of the recent un-/semi-structured pruning approaches for decoder-only LLMs. This comparison is unusual because these two are in entirely different classes of pruning techniques. However, we posit LayerChop against 2:4 semi-structured pruning methods, Wanda~\citep{wanda} and sparseGPT~\citep{sparseGPT}, because 2:4 pruned models improves speed on NVIDIA hardware using specialized kernels. We show that with continued pretraining LayerChop outperforms 2:4 semi-structured pruning on llama2 from Wanda and sparseGPT. Additionally, Table~\ref{tab:speedup} highlights the end-to-end inference speed gained by these methods. LayerChop incurs additional training costs due to continued pretraining but makes up the difference by being 1.84x faster than the base 7B llama2 model against 1.24x end-to-end speed achieved by the 2:4 pruned model. Given a training token budget for LayerChop, we can calculate the number of inference queries it will take to break even compared to the 2:4 pruning approach, which is one-shot pruning.

Table~\ref{tab:structured-comparison} highlights how LayerChop compares to some of the recently proposed structured pruning methods for larger decoder-LLMs. Shortened Llama~\citep{shortenedLlama} is a layer pruning method that trains LoRA~\citep{lora} weights after pruning, while SliceGPT~\citep{sliceGPT} and LLMSurgeon~\citep{llmSurgeon} are dimension pruning approaches that prune model dimensions based on a heuristic. Since the base model and the evaluation setup are different across our work and these papers, we highlight the overlapping number of tasks where results were presented in each of these works, and compute the average of the reported accuracies for the base model and the final pruned models only on the tasks that overlap. We propose a percentage decrease in average task accuracy to measure each pruning strategy's effectiveness. As we can see, LayerChop consistently outperforms all other heuristic-based structured pruning approaches while pruning $50\%$ of the model, a pruning rate which is equaled only by LLMSurgeon~\citep{llmSurgeon}.

\begin{table}[!b]
\centering
\footnotesize
\begin{tabular}{@{}c|cc@{}}
\toprule
compression & \textbf{2:4} & \textbf{LayerCHOP} \\ \midrule
\textbf{Uncompressed latency} & 312ms & 351ms \\
\textbf{Compressed latency} & 251ms & 191ms \\
\textbf{Speed} & 1.24x & 1.84x \\
\textbf{Breakeven query} & 240M & 360M \\ \bottomrule
\end{tabular}
\caption{Comparison between end-to-end inference speeds of 2:4 semi-structured pruned models and LayerCHOP. The numbers for 2:4 pruning are taken from Wanda~\citep{wanda}. LayerCHOP, a form of structured pruning, is 1.84x faster than the base model, against 1.24x speed improvement of 2:4 pruned model. The additional cost of training LayerCHOP can be subsumed by the gain in inference speed in the breakeven query number given for each type of pruning.}
\label{tab:speedup}
\end{table}

\subsection{Comparisons against simple pretraining baselines}
\label{sec:results-pretraining}

\begin{figure*}[!t]
\centering
\makebox[\textwidth][c]{
\includegraphics[width=\textwidth]{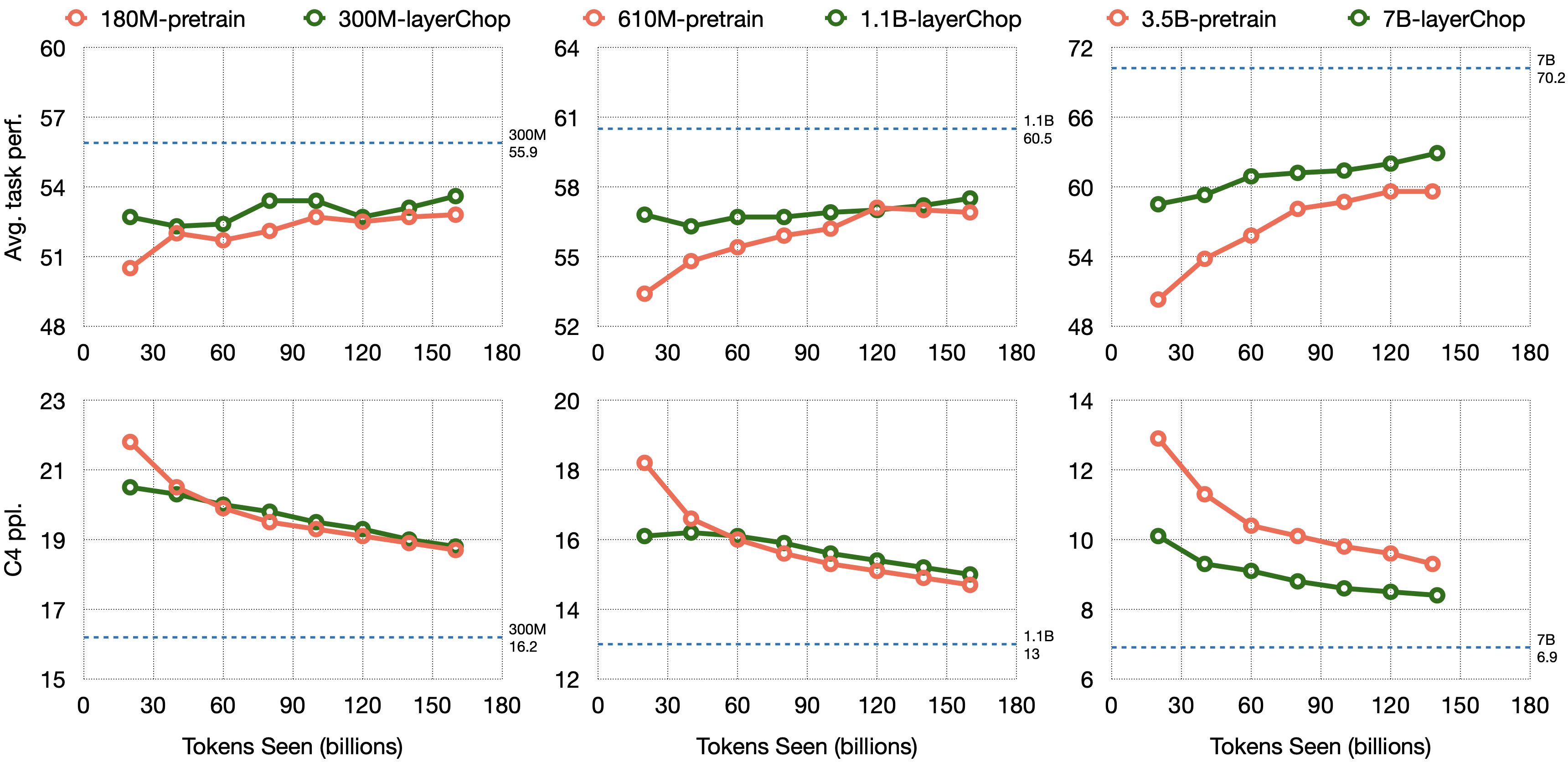}
}
\caption{Language modeling and average task accuracy performance plotted against token budget for LayerCHOP against pretraining baselines. We observe that for smaller 300M and 1.1B models, pruning and finetuning of models converge to pretraining baselines. However, this trend is not obvious at the 7B scale.}
\label{fig:main-figure}
\end{figure*}

We take the best configuration (Appendix~\ref{sec:appendix-layer-chop}) of our most competitive pruning strategy, i.e., LayerCHOP, and continue pretraining for an extended token budget of 160B tokens from C4 corpus to compare against pretraining a similar-sized model from scratch at 3 models scales: 300M, 1.1B, and 7B. We present the average task accuracies over the evaluation suite defined in Section~\ref{sec:eval-setup} and language modeling perplexity on the C4 eval set defined in Section~\ref{sec:experimental-setup} in Figure~\ref{fig:main-figure}.

As we see in both the task average and perplexity trends, pruning with continued training eventually converges towards pretraining at the 300M and 1.1B scale after a certain number of tokens. At the same time, it looks to move towards convergence at the 7B scale. For coarse-grained model pruning, this trend deviates from the idea of ``train large and then compress'', as we show in these experiments that if the token budget is large enough, it might be a better idea to pretrain models from scratch instead of pruning them in a coarse fashion. Note that this trend contrasts what pretraining with fine-grained pruning observes in Sheared Llama~\citep{sheared-llama}.

\begin{figure}[!b]
\centering
\makebox[\columnwidth][c]{
\includegraphics[width=0.7\columnwidth]{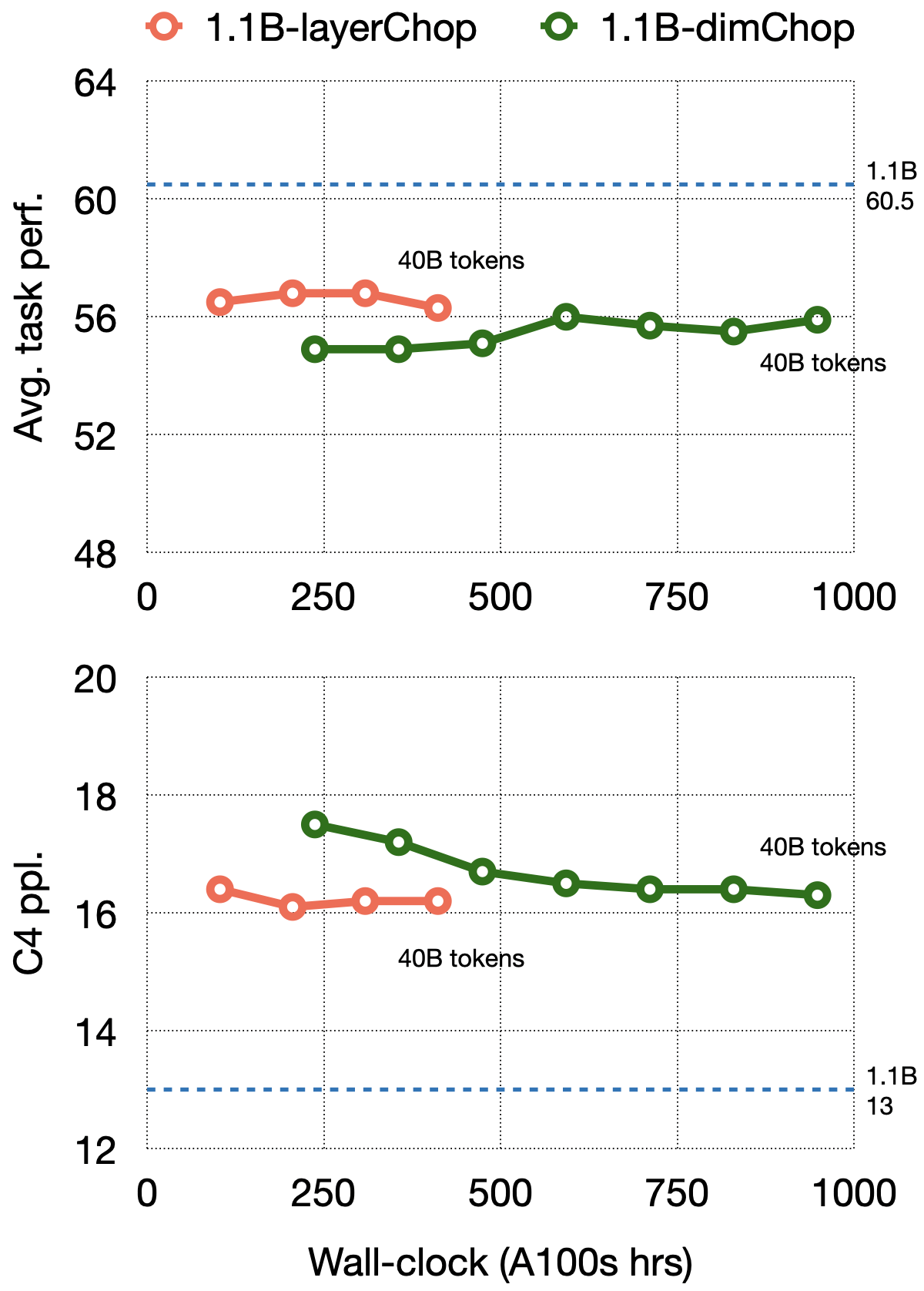}
}
\caption{Average task accuracy and language modeling comparison between LayerCHOP and DimCHOP on a fixed compute budget. We demonstrate that LayerCHOP is 2$\times$ more efficient during the compression process than DimCHOP by pruning all layers immediately (Appendix~\ref{sec:appendix-layer-chop}), while DimCHOP opts for a more iterative pruning approach.}
\label{fig:coarse-vs-fine}
\end{figure}

\subsection{Comparing LayerCHOP and DimCHOP}

In this section, we compare the results of LayerCHOP and DimCHOP in terms of performance and efficiency. Figure~\ref{fig:coarse-vs-fine} shows how both methods perform in the same ballpark on average task performance and language modeling after consuming the same number of pretraining tokens; however, the algorithmic differences in the two pruning strategies make LayerCHOP much more efficient to train compared to DimCHOP. LayerCHOP instantly prunes half the number of layers from a model and continues pretraining only half-sized models. At the same time, DimCHOP is an iterative algorithm that prunes dimensions after every few steps. However, it further trains the zeroed-out dimensions before re-evaluating them for pruning based on importance scores (Section~\ref{sec:dim-chop}).

\begin{figure*}[t]
\centering
\makebox[\textwidth][c]{
\includegraphics[width=0.7\textwidth]{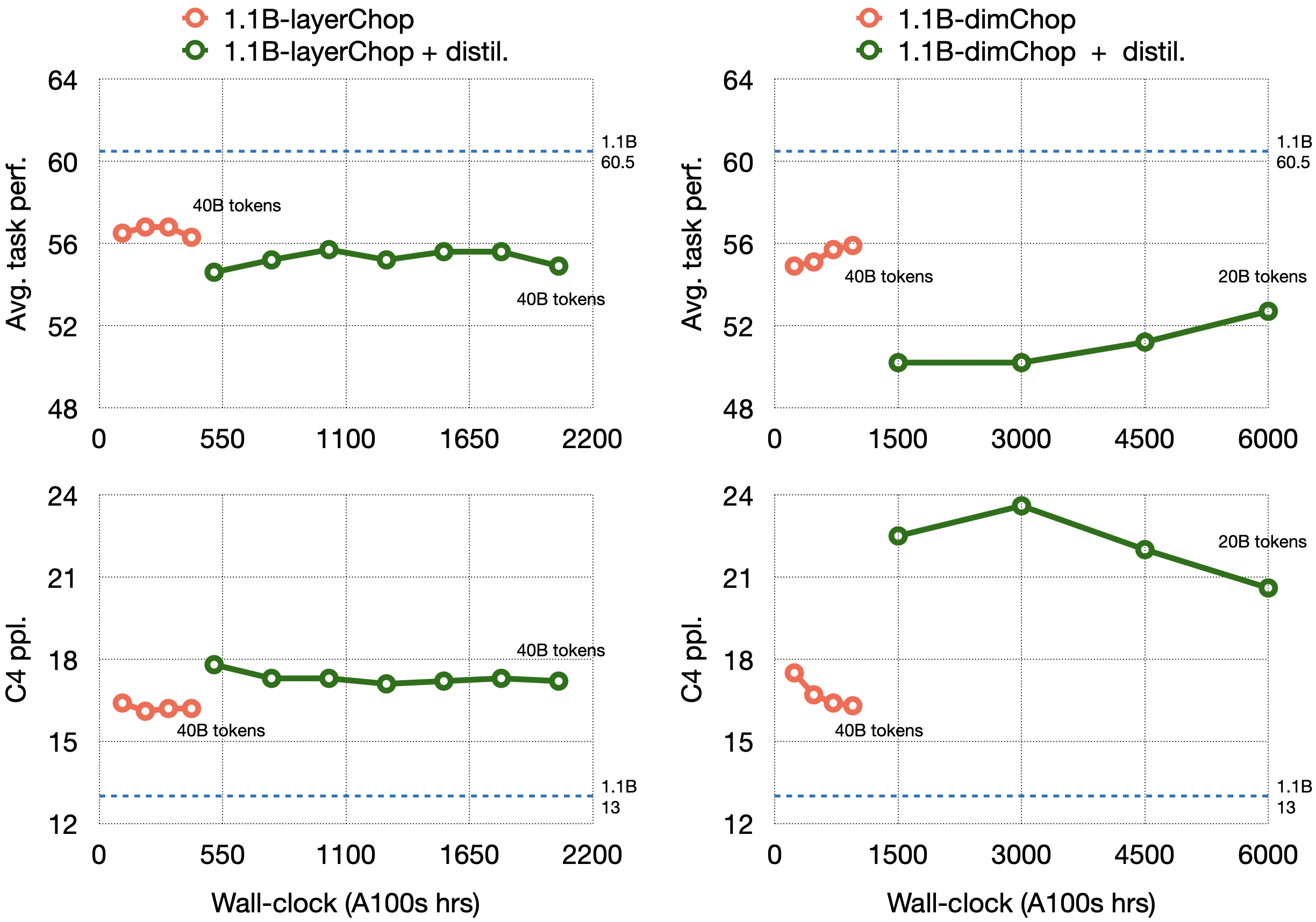}
}
\caption{Average task accuracy and language modeling comparison between LayerCHOP, DimCHOP and their distillation loss augmented versions. In a large data regime, for large student and teacher models, we see that adding distillation become inefficient with our pretrain-then-finetune paradigm for LLM compression.}
\label{fig:distillation-comparison}
\end{figure*}

\subsection{Efficiency of distillation with a large pretraining corpus}

Figure~\ref{fig:distillation-comparison} compares average task performance and language modeling perplexity results for LayerCHOP and DimCHOP against distillation augmented versions of the algorithms described in Section~\ref{sec:distil-chop}. Homotopic distillation~\citep{homotopic-distill} finds this setup to be efficient and highly performant for task-agnostic compression of BERT-style models; however, in our large data regime with bigger decoder-only LLMs, we find that augmenting pruning with distillation losses not only underperforms on average task score, when evaluated in a zero-shot manner but also on language modeling.

With a larger teacher and student model, augmenting a pruning setup distillation losses in different parts of the architecture incurs a heavy efficiency penalty, as is evident from Figure~\ref{fig:distillation-comparison}. Previous distillation approaches are also orthogonal to newer efficient modeling paradigms like flash-attention, which does not materialize a $O(n^2)$ attention matrix to which we can apply the earlier proposed distillation objective in Section~\ref{sec:distil-chop}.


\section{Discussion and Conclusion}
\label{sec:discussion}

This work discusses how existing LLM compression methods do not follow the pretrain-then-finetune paradigm. We show that with additional continued pretraining of compressed models after pruning, coupled with embarrassingly simple pruning techniques, can beat complex structured pruning approaches and compete with semi-structured compression methods with improved efficiency. Our LayerCHOP methods set new performance standards for LLM compression while improving inference speeds by 1.84$\times$ over the baseline model, compared to the limited 1.24$\times$ end-to-end speed improvement from 2:4 semi-structured compression via specialized NVIDIA accelerator kernels.

The work closest to ours is Sheared Llama~\citep{sheared-llama}, where they apply Co-Fi~\citep{co-fi} like fine-grained pruning and continued pretraining on RedPajama data. Their work selects components of a 7B model (e.g., attention heads) to assemble smaller models, at the 1.3B and 2.7B scales, using a pruning heuristic that trains further on domain-specific data. Our work differs by recovering state-of-the-art compression results by continuing to pretrain despite pruning models at a much coarser granularity. Our methods also require no domain-specific data selection, as we continue pretraining on same pretraining data distribution. 

Finally, we demonstrate that under this new pretrain-then-finetune paradigm for LLM compression, augmenting pruning with distillation losses does not improve performance as is the case for BERT-style models. We also identify how student-teacher distillation on large pretraining corpus results in inefficient compression algorithms. CHOP offers compression compatible with contemporary modelling optimizations (e.g., Flash Attention) to surpass distillation with embarassingly simple techniques. We hope our work contributes to a new conversation on how to train a compressed large language model, and how to learn from large corpora for better, smaller models. 


\section*{Limitations}


While our work is in the spirit of reducing model size and improving efficiency --- we require significant computational resources for our experiments demanding both high energy usage and processing power. Experiments such as the model pretraining and distillation at scale demand multiple days of training time using 32xA100 GPUs with a high bandwidth interconnect. Therefore, reproducing our experiments are only reasonably tractable with commensurate GPU resources which may be infeasible for some researchers.

Additionally, we demonstrate our findings compared to pretraining and distillation approaches and recently published alternatives in our decoder-only setup. We take this approach to report how the most typical compression strategy can be ported to a contemporary LLM. Our finidings do not indiciate distillation to be a potentially efficient compression strategy for GPT-style models in a large data regime, however, our work is limited in that there may exist \emph{some atypical} distillation strategy with even better performance. We encourage future work and discussion of how these methods can be improved in this regard.

\section*{Ethics Statement}
We report all pretraining experiments with the widely used C4 corpus. This corpus has been found to contain harmful artifacts and biases \citep{dodge-etal-2021-documenting} which our models may inherit, however, the study of this phenomena is outside of the scope of our work but may inform future study. Model compression has been linked to increased bias and toxicity in a model \citep{hooker2020characterising} but it is currently unclear how such effects extend to our setting; particularly as we expose the student to the same corpus as the teacher. Further study is needed in this area to examine how compression influences biases in increasingly large language models \citep{solaiman2023evaluating}.

\bibliography{anthology,custom}

\newpage
\appendix

\section{Model Architecture}
\label{sec:appendix-arch}

\begin{table}[!t]
\centering
\adjustbox{max width=\columnwidth}{
    \begin{tabular}{@{}ccccccc@{}}
    \toprule
    {\# Params} & {Dim} & {Heads} & {Layers}  & {Batch Size} & {LR} & {Token Budget}\\
    \midrule
    180M & 1024 & 16 & \textbf{12} & 2M & 6.0e-4 & ~160B\\
    300M & 1024 & 16 & 24 & 2M & 3.0e-4 & ~160B\\ \midrule
    610M & 2048 & 16 & \textbf{12} & 2M & 2.5e-4 & ~160B\\
    1.1B & 2048 & 16 & 24 & 2M & 2.0e-4 & ~160B\\ \midrule
    3.5B & 4096 & 32 & \textbf{16} & 4M & 3.0e-4 & ~160B\\
    7B & 4086 & 32 & 32 & 4M & 3.0e-4 & ~2T\\ \midrule
    \bottomrule
    \end{tabular}
}
\caption{Configurations for models used in our experiments at different scale. The 7B model used is pretrained llama2-7B.}
\label{table:pretrained-model-configs}
\end{table}

We list details about our decoder-only models in Table.~\ref{table:pretrained-model-configs}. For the 7B size, we prune a pretrained llama2-7B~\citep{llama2}. For smaller model sizes, we train a custom architecture model on 160B C4 tokens. 

For the custom model, we follow the PaLM architecture \citep{palm} owing to improved throughput efficiency. Specifically, the attention and feed-forward network (FFN) modules are parallel instead of sequential \citep{gpt-2}. SwiGLU activation \citep{swiglu} is used in the FFN module. Multi-head attention uses the equivalent Flash-Attention \citep[FA]{flash-attn} implementation. The first layer of the FFN module and the layers generating attention query, key, and value are fused. Similarly, the second layer of the FFN module and the feed-forward layer after the attention operation are fused. The LayerNorm \citep{layernorm} is before the first fused feed-forward layer. The query and the key vectors are passed through additional layer normalization layers for increased training stability following \citet{vit-22b}. This block structure is repeated with skip connections to form our decoder-only Transformer architecture.

\section{Evaluation setup}
\label{sec:appendix-eval}

We detail the tasks used in our zero-shot evaluation suite in Table~\ref{table:end-tasks}. Each task in the table reports either classification accuracy or length normalized classification accuracy. Our evaluation suite, which is online (runs as a validation loop after some training steps), is adapted to match the results from Eleuther AI eval harness~\citep{eai-eval-harness}.

\begin{table}[!t]
\resizebox{\columnwidth}{!}{%
\begin{tabular}{@{}lll@{}}
\toprule
\textbf{category} & \textbf{task} & \textbf{metric} \\ \midrule
\multirow{4}{*}{\begin{tabular}[c]{@{}l@{}}common sense\\ reasoning\end{tabular}} & PIQA \citep{piqa} & len norm acc \\
 & Hellaswag \citep{hellaswag} & len norm acc \\
 & Winogrande \citep{winogrande} & acc \\ \midrule
\multirow{4}{*}{\begin{tabular}[c]{@{}l@{}}science question\\ answering\end{tabular}} & OpenBookQA \citep{obqa} &  len norm acc \\
 & SciQ \citep{sciq} & acc \\
 & Arc-Easy \citep{arc} & acc \\ \bottomrule
\end{tabular}%
}
\caption{Zero-shot downstream tasks for evaluating our compressed models and baselines. Each task either reports classification accuracy or length normalized classification accuracy.}
\label{table:end-tasks}
\end{table}

\section{LayerCHOP}
\label{sec:appendix-layer-chop}

\begin{figure}[!b]
\centering
\makebox[\columnwidth][c]{
\includegraphics[width=\columnwidth]{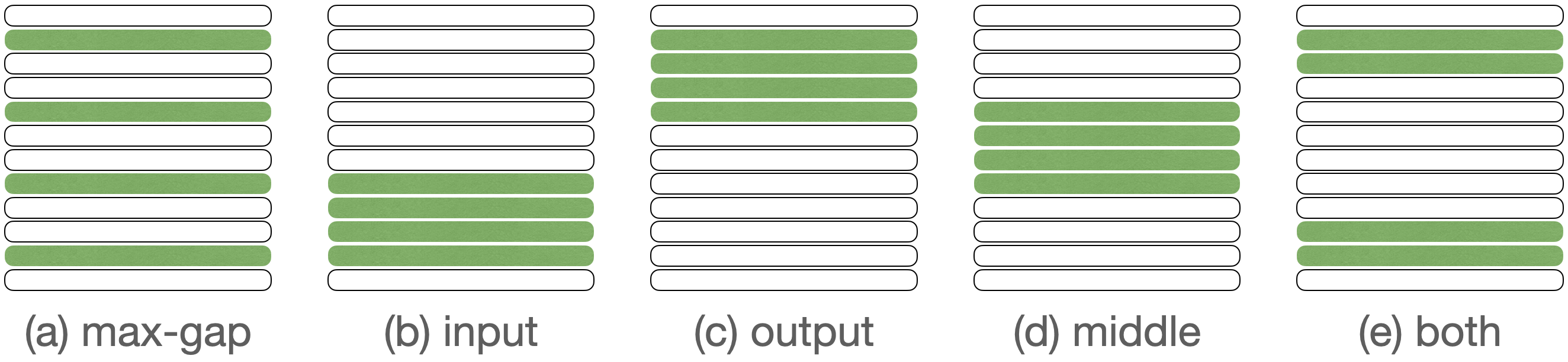}
}
\caption{Truncated initialization configurations for layer pruning in a decoder-only language model. Highlighted layers (\textcolor{darkolivegreen}{green}) are removed. We retain the first and last layer as these layers interact with the embedding table.}
\label{fig:pruning-locations}
\end{figure}

Our baseline models contain 24 decoder layers. To determine which layers we should prune for the best compression performance, we define five layer pruning configurations shown in Figure~\ref{fig:pruning-locations}, each removing 12 out of the 24 decoder layers of the 300M and 1.1B models. In all these pruning configurations, we always keep the first and the last layers because they interact with the embedding table. We made this design choice based on early experiments. Table~\ref{table:pruning-locations} summarizes the results of this ablation. We report the perplexity score on the C4 validation set and the average task accuracy across 6 tasks in our evaluation suite (Table \ref{table:end-tasks}).

For the base 300M model, pruning configurations of \emph{max-gap} and \emph{both} perform the best out of the five possible configurations. For the 1.1B model, pruning layers from the \emph{input} configuration yielded the best results for both reported metrics. The \emph{output} pruning configuration resulted in the worst performance across model sizes, suggesting that pruning layers towards the output side of the model should be avoided. Given these results, we use the pruning configuration of \emph{max-gap} for all our 300M model experiments and the configuration of \emph{input} for all our 1.1B model experiments, and use the configuration from 1.1B model experiments for our llama2-7B pruning experiments.

\begin{table}[!t]
\resizebox{\columnwidth}{!}{%
\begin{tabular}{@{}cccccccc
>{\columncolor[HTML]{C0C0C0}}c @{}}
\toprule
Model & Token Budget & Task Metric & max-gap & input & output & middle & both & Pre-compression \\ \midrule
300M-160B & 20B & ppl $\left(\downarrow\right)$ & \textbf{23.0} & 24.5 & 25.6 & 24.0 & \textbf{23.0} & 16.2 \\
300M-160B & 20B & avg acc $\left(\uparrow\right)$ & \textbf{53.2} & 52.9 & 51.6 & 52.9 & \textbf{53.2} & 55.8 \\
\hline
1.1B-160B & 20B & ppl $\left(\downarrow\right)$ & 18.1 & \textbf{17.3} & 22.0 & 18.6 & 18.6 & \cellcolor[HTML]{C0C0C0}13.0 \\
1.1B-160B & 20B & avg acc $\left(\uparrow\right)$ & 54.8 & \textbf{55.1} & 53.1 & 54.7 & 53.8 & 59.8 \\ \bottomrule
\end{tabular}
}
\caption{Influence on average task performance and C4 validation perplexity of different truncated initialization strategies from Figure \ref{fig:pruning-locations} for models of size 300M and 1.1B. The average task performance score is across 6 tasks listed in Table \ref{table:end-tasks}, and higher numbers are better. For perplexity scores, lower is better.}
\label{table:pruning-locations}
\end{table}

\begin{figure}[!b]
\centering
\makebox[\columnwidth][c]{
\includegraphics[width=\columnwidth]{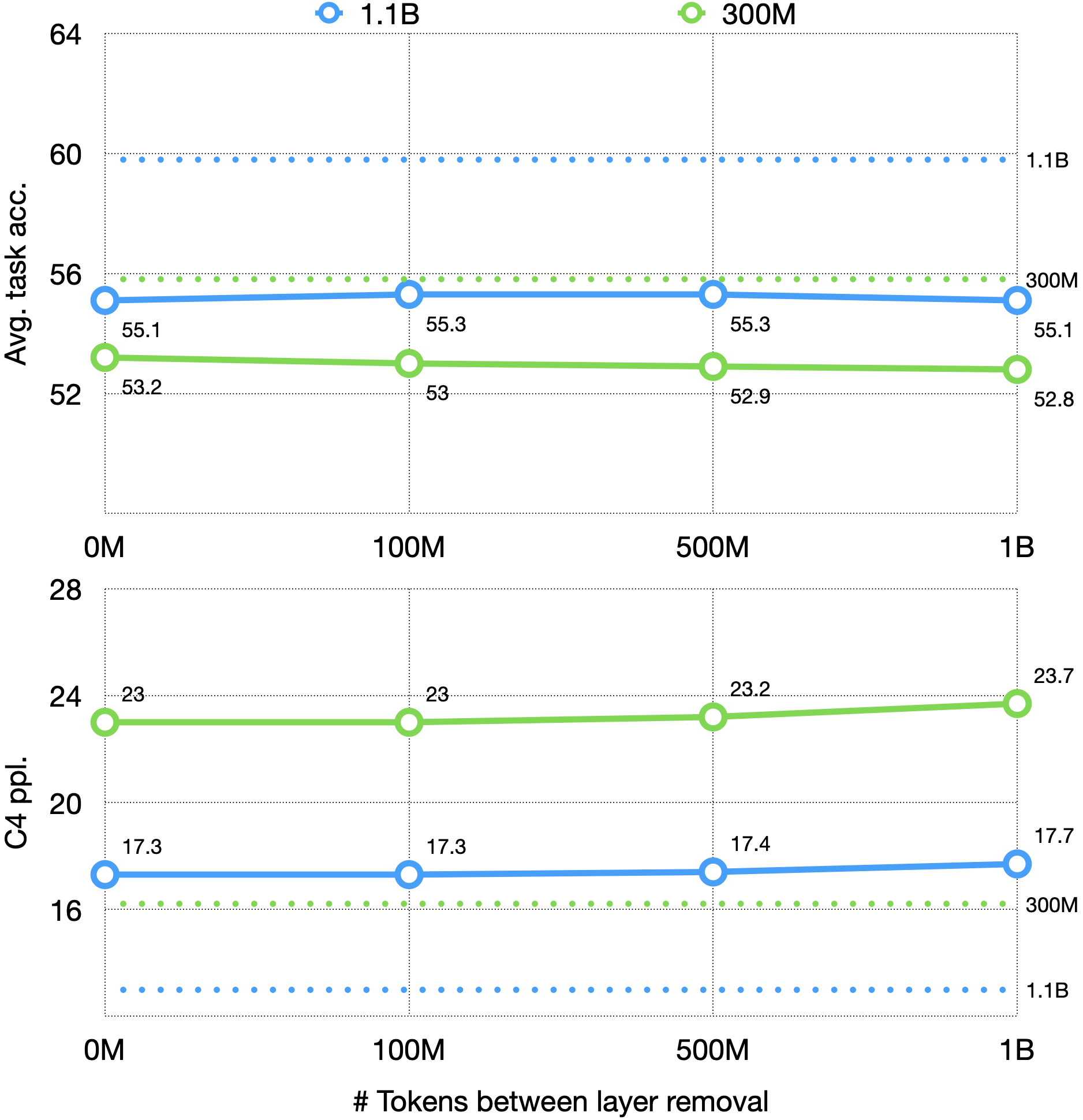}
}
\caption{Average task accuracy (Table \ref{table:end-tasks}) and perplexity on the C4 validation set for model sizes 300M and 1.1B comparing schedules for \emph{when} to prune layers during continued pretraining. We find a marginal performance degradation as we remove layers one by one further apart during continued pretraining.}
\label{fig:pruning-gap}
\end{figure}

We can decide when to prune layers while training models with the LM loss in one of two ways: either remove selected layers at once or remove them one by one, each after a fixed number of training tokens. We run this experiment in four configurations to see if increasing the gap between each layer pruning increases training stability or model performance. The four configurations are: dropping all layers at once (0M token gap between pruning each layer) or pruning them after 100M, 500M, and 1B training tokens each. We run this experiment for the 300M and 1.1B model sizes. We prune 12/24 layers from our decoder-only models for this ablation, each at a training token gap mentioned in one of the four configurations above. The result of this experiment is summarized in Figure~\ref{fig:pruning-gap}. Pruning layers one by one with an increasing token budget between each layer pruning does not benefit the average task accuracy or C4 validation perplexity. In fact, there is a marginal preference to prune layers as early into the training as possible. Hence, we decided to prune all 12/24 layers simultaneously for other experiments.

\end{document}